# View subspaces for indexing and retrieval of 3D models

Helin Dutagaci[a], Afzal Godil[a], Bülent Sankur[b], Yücel Yemez[c]
[a]National Institute of Standards and Technology, Gaithersburg, MD, USA;
[b]Dept. of Electrical-Electronics Engineering, Boğaziçi University, Istanbul, TURKEY;
[c]Dept. of Computer Engineering, Koç University, Istanbul, TURKEY

**ABSTRACT**

View-based indexing schemes for 3D object retrieval is gaining popularity since they provide good retrieval results. These schemes are coherent with the theory that humans recognize objects based on their 2D appearances. The view-based techniques also allow users to search with various queries such as binary images, range images and even 2D sketches.

The previous view-based techniques use classical 2D shape descriptors such as Fourier invariants, Zernike moments, SIFT-based local features and 2D DFT coefficients. These methods describe each object independent of others. In this work, we explore data driven subspace models, such as PCA, ICA and NMF to describe the shape information of the views. We treat the depth images obtained from various points of the view sphere as 2D intensity images and train a subspace to extract the inherent structure of the views within a database.

## 1. INTRODUCTION

Shape-based indexing and retrieval of 3D objects became a necessity with the increase of the number and diversity of 3D model databases, ranging from multimedia, archeological and biomedical applications to CAD/CAM-based manufacturing. Among the current approaches proposed for 3D object retrieval tasks, view-based indexing schemes are gaining popularity since they provide good retrieval results. These schemes are coherent with the theory that humans recognize objects based on their 2D appearances. The view-based techniques also allow users to search with various queries such as binary images, range images and even 2D sketches.

The state-of-the-art view-based techniques use classical 2D shape descriptors. Some examples are Fourier invariants[1], Zernike moments[1], 2D Discrete Fourier Transform (DFT) coefficients[2], and Scale Invariant Feature Transform (SIFT)-based local features[3]. In this work, we evaluate the retrieval performance of three data driven subspace models: Principal Component Analysis (PCA), Independent Component Analysis (ICA) and Nonnegative Matrix Factorization (NMF).

One of the disadvantages of view-based methods is the theoretically infinite number of possible views. One solution is to adequately sample the view sphere. In general, the same number of views is used for all the models in the database. In this work, we make this number variable and dependent on the object characteristics. While applying pose normalization to the object via PCA, we assess the eigenvalues and their ratios to decide on the number of views. For objects that yield very distinct eigenvalues, we rely on the six canonical views obtained by PCA. For others with close eigenvalues, i.e. spherical kind of objects, we sample the view sphere with more points.

### 1.1 Related work

In view-based shape description approach, 2D images of the model are rendered from different view-points. Then, each of the 2D image is described with a set of local or global 2D features. Two objects are considered to be similar if the distance between the descriptors of the views is low. The state-of-the-art view-based approaches differ in the following ways: 1) The sampling scheme of the view sphere; for example six canonical views or vertices of a dodecahedron or geodesic sphere; 2) Representation of the views, such as depth images versus silhouettes or contours; 3) Feature extraction methods, such as global Fourier or Zernike descriptors, or local SIFT descriptors.

One of the first view-based method, known as the Light Field Descriptor (LFD), was proposed by Chen et al.[1] They sampled the view sphere using the vertices of a dodecahedron and represented the views as silhouettes, resulting in 10 silhouettes per model. They encoded a silhouette with a combination of Fourier and Zernike descriptors. The dissimilarity between two models was calculated over all the possible rotations of the dodecahedron. They also provided techniques to reduce the computational cost. Vranic[2] used Fourier descriptors extracted from both the depth images and silhouettes. He selected the six canonical views obtained through CPCA-based pose normalization of an object. Furuya and Obhuchi[3], rendered a 3D model from 42 viewpoints placed uniformly on the view sphere. They utilized Scale Invariant Feature Transform (SIFT) in order to extract local descriptors from the depth images. Then these SIFT features are encoded with a visual codebook and the depth images were described with a histogram of the encoded features.

Ansary et al.[4] used a Bayesian approach to select the characteristic views of a model. In another work[5], Curvature Scale Space is used to describe the views. Chaouch et al.[6] proposed to associate a relevance index to each view in order to increase the impact of more informative views. In a recent paper of Chaouch and Verroust-Blondet[7], 20 depth images were rendered from the vertices of a dodecahedron. Then the depth images were represented by a set of depth lines quantized with respect to the gradient. The encoded depth lines of two models were compared via Dynamic Programming.

### 1.2 Contributions and outline

Our contribution to the state-of-the art of view based methods is the use of subspace-based features to describe the object views rendered as depth images. This approach was proposed by Murase and Nayar[8] for recognition of objects from 2D intensity images; but to the best of our knowledge, it was not considered for the geometric projection of the 3D models in the context of 3D model retrieval.

Another contribution of this work is the pre-filtering of the database with respect to the elongation characteristics of the query model. For example, the dissimilarity between two models based on the global descriptors will be high if one of them has a sphere-like structure and the other is highly elongated. The filtering is extremely simple, since we just use the eigenvalues derived from the PCA-based pose normalization of the models.

The paper is organized as follows: In Section 2, we introduce the pose-normalization, the categorization and view generation steps. In Section 3, we describe the subspace-based feature extraction schemes. In Section 4, we clarify how the matching is performed between two models. We give experimental results in Section 5, and conclude in Section 6.

## 2. VIEW SAMPLING

In this section, we first describe the pose normalization stage. Then, we introduce the categorization step which is guided by the eigenvalues derived from the pose normalization. Finally, we describe the view generation based on the geodesic sphere.

### 2.1 Pose normalization

We rotate a 3D triangular mesh into a reference frame via CPCA pose normalization, which was proposed by Vranic[9]. First, a $3 \times 3$ covariance matrix is calculated over the point set of all triangles:

$$\mathbf{C} = \frac{1}{S} \int_I \mathbf{p} \cdot \mathbf{p}^T dp$$

where $S$ is the total surface area, $\mathbf{p}$ is the 3D coordinate vector of a point on the surface, and $I$ is the set of points on the object surface. The eigenvalues of $\mathbf{C}$ are calculated and are sorted in decreasing order such as $\{\lambda_1 \geq \lambda_2 \geq \lambda_3\}$. The corresponding eigenvectors define the principal axes of the object, $\mathbf{e}_1$ defining the first principal axis, $\mathbf{e}_2$ the second, and $\mathbf{e}_3$ the third. The object is rotated such that $\mathbf{e}_1$, $\mathbf{e}_2$, and $\mathbf{e}_3$ coincide with $x$, $y$, and $z$ axes respectively.

## 2.2 Categorization with respect to eigenvalues

The CPCA-based pose normalization gives robust results if the model is well-elongated, i.e. it has distinct eigenvalues. The three eigenvalues indicate the extent of the model in the corresponding orientations; hence their ratios give an indication of the elongation of the model along the corresponding plane. Let us define the following ratios:

$$a_1 = \frac{\lambda_2}{\lambda_1}; \qquad a_2 = \frac{\lambda_3}{\lambda_1}; \qquad a_3 = \frac{\lambda_3}{\lambda_2}$$

The values of $a_1$, $a_2$, and $a_3$ range between 0 and 1, since we have sorted the eigenvalues with decreasing order. If $a_1$ is close to 1, i.e. $\lambda_1$ and $\lambda_2$ are close to each other, then the normalized object is equally extended in $x$ and $y$ directions. The object is not elongated and the first and second principal directions are ambiguous. On the other hand, a small value of $a_1$ indicates a well elongated and well directed object along the $x$ direction. The same comments can be made for $a_2$ and $a_3$.

These ratios give a rough description of the global geometry of the object. For certain classes of objects, these ratios more or less remain stable. In an object retrieval application, filtering the database with respect to the eigenvalue ratios of the query model can greatly reduce the search space. The ratios can be used to roughly categorize the objects into sphere-like objects, cylindrical objects, planar objects, etc. Figure 1 gives an idea of the types of objects with respect to the eigenvalue ratios, $a_1$ and $a_3$. Notice that $a_2$ completely depends on $a_1$ and $a_3$.

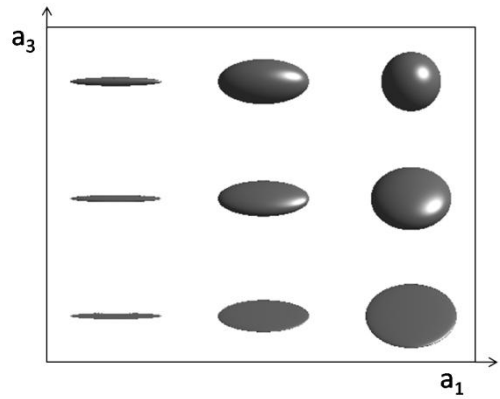

Figure 1. Visualization of the elongation properties of 3D models with respect to the eigenvalue ratios, $a_1$ and $a_3$.

## 2.3 Views on geodesic sphere

We follow the view sampling method proposed by Lian et al.[10], where the authors use the vertices of a geodesic sphere surrounding the model in order to obtain views. Figure 2 shows some examples of geodesic spheres. We subdivide a regular octahedron to obtain finer geodesic spheres with more vertices. In this work, we have only used two levels of geodesic spheres; the first one is the original octahedron and the second one is the subdivided version with 16 vertices. We capture the depth images as seen from the vertices of these two spheres. The depth images are mapped onto a regular grid of resolution $128 \times 128$. Figure 2 shows two examples of the view extraction process.

The number of views for an object is determined with respect to the eigenvalue ratios, $a_1$ and $a_3$. When these values are low, we assume that the object is well elongated along the principal axes, and that the six canonical views are enough to describe the object. These six canonical views correspond to the vertices of the regular octahedron. For models with high $a_1$ or $a_3$, we render the model from the 16 vertices of the geodesic sphere, which was obtained via subdivision of the regular octahedron (Figure 3). The details are described in Section 4.

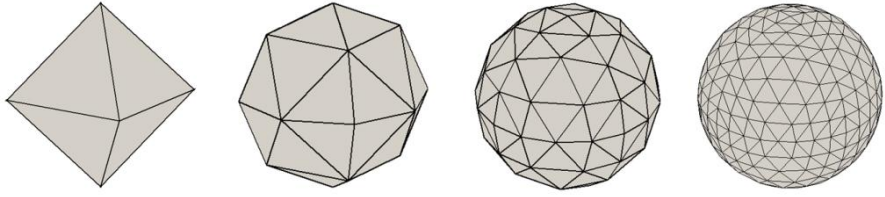

Figure 2. Examples of geodesic spheres

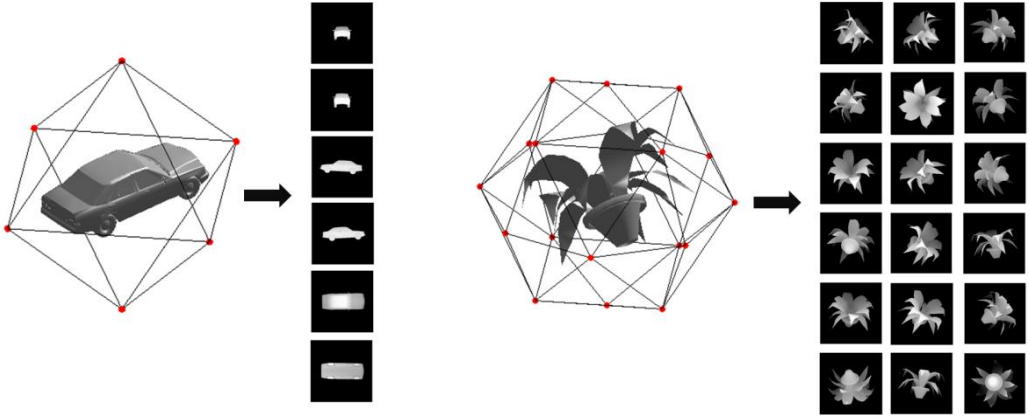

Figure 3. Two examples of view sampling. The object on the left has distinct eigenvalues, and it is well aligned with CPCA normalization. Therefore we capture only the six canonical views. The object on the right is highly spherical, so we capture more views.

## 3. SUBSPACE FEATURES

Subspace methods find a set of vectors that describe the significant statistical variations among the observations. These vectors form the basis of a subspace where most of the meaningful information for certain class of processes is preserved. These methods have the additional advantage of greatly reducing the dimensionality of the observations. In this specific application, our observations are the depth images generated from the complete 3D models. We construct a data matrix $\mathbf{X}$, by collecting $N$ depth images from a training set, converting them to one dimensional vectors of length $M$, and putting these vectors $\{\mathbf{x_1}, \mathbf{x_2}, \ldots \mathbf{x}_N\}$ into the columns of the data matrix $\mathbf{X}$. Then, we analyze this matrix via one of the following techniques: Principal Component Analysis, Independent Component Analysis, and Nonnegative Matrix Factorization. Once basis vectors of the subspace are formed, any new observation, i.e. a new depth image, is projected onto the subspace and the projection coefficients are used as the descriptors of that depth image.

### 3.1 Principal Component Analysis

PCA is an analysis technique that is based on the decorrelation of the data using second order statistics. The eigenvectors of the $M \times M$ covariance matrix, $\mathbf{G} = (\mathbf{X} - \bar{\mathbf{x}})(\mathbf{X} - \bar{\mathbf{x}})^\mathbf{T}$ gives the principal directions of variations. Here, $\bar{\mathbf{x}}$ denotes the mean of the training vectors. Let $\{\mathbf{v_1}, \mathbf{v_2}, \ldots \mathbf{v_K}\}$ be the first $K$ eigenvectors of $\mathbf{G}$ with corresponding eigenvalues $\{\alpha_1 \geq \alpha_2 \geq \ldots \geq \alpha_K\}$. These vectors model the largest variations among the training samples, therefore are considered to capture most of the significant information. The amount of information maintained depends on $K$ and the spread of eigenvalues. The projection of an input vector $\mathbf{x}$ onto the PCA subspace is given by $\mathbf{b} = \mathbf{V}^T\mathbf{x}$, where $\mathbf{V}$ represents the $M \times K$ projection matrix formed as $[\mathbf{v_1}\ \mathbf{v_2}\ ...\ \mathbf{v_K}]$.

The data matrix $\mathbf{X}$ is formed by collecting six canonical views of each of the 907 training models in the Princeton Shape Benchmark (PSB)[11], so there are $6 \times 907$ observations to be analyzed. Figure 4 shows the first four principal modes of variations among the depth images of PSB training models.

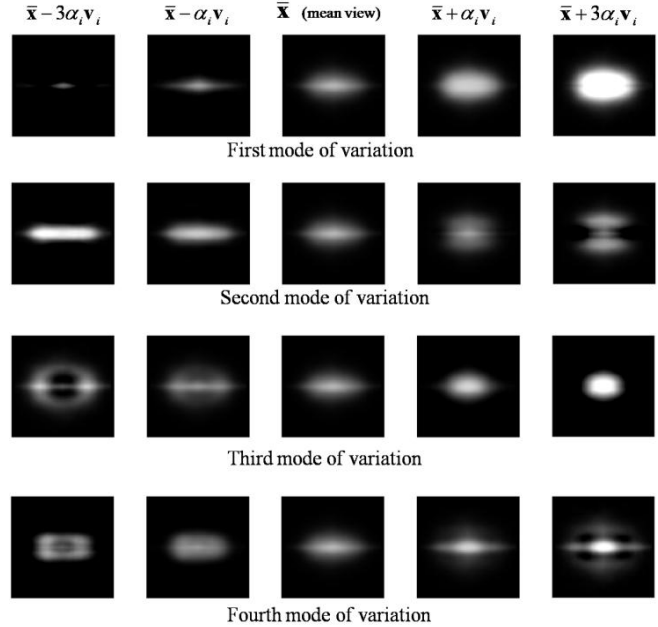

Figure 4. Modes of variations of training depth images through PCA

### 3.2 Independent Component Analysis

ICA is a generalization of PCA in that it removes correlations of higher order statistics from the data. With ICA, we assume that the observed signals $\{\mathbf{x_1}, \mathbf{x_2}, ...\mathbf{x}_N\}$ result from linear mixtures of $K$ source signals $\{\mathbf{s_1}, \mathbf{s_2}, ...\mathbf{s_K}\}$. We admit the signal model $\mathbf{X} = \mathbf{AS}$, where $\mathbf{A}$ is the matrix of mixing coefficients and $\mathbf{S}$ contains the sources in its rows. Both the source signals and the mixing coefficients are unknown, and need to be estimated. Our aim is to find a linear transformation $\mathbf{W}$, such that $\hat{\mathbf{S}} = \mathbf{WX}$, where $\mathbf{W}$ is the separating or de-mixing matrix. We estimate $\mathbf{W}$ using the FastICA algorithm[12]. Figure 5 shows the ICA basis images obtained from analysis of the depth images of PSB training set.

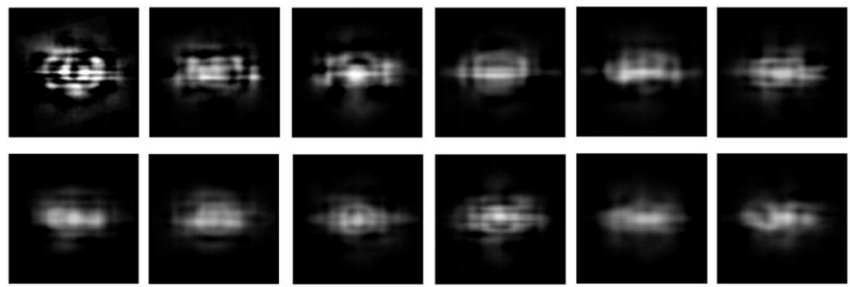

Figure 5. ICA basis for depth images

## 3.3 Nonnegative Matrix Factorization

Given a nonnegative data matrix, $\mathbf{X}$, of size $M \times N$, we factorize it into two nonnegative matrices $\mathbf{U}$ and $\mathbf{H}$, such that $\mathbf{X} \approx \mathbf{UH}$, with. sizes $M \times K$ and $K \times N$, respectively. $\mathbf{U}$ contains the basis vectors in its columns and $\mathbf{H}$ is constituted of combination coefficients. We use the update rules described by Lee and Seung[13] to estimate the nonnegative factors. The objective function is taken as $\|\mathbf{X} - \mathbf{UH}\|^2$, where $\|\bullet\|$ is the Frobenius norm and the factor matrices are constrained to have nonnegative elements[13].

Figure 6 shows the NMF basis images obtained from analysis of the depth images of PSB training set. Notice that each basis image is only active on a very localized region, which is in contrast with the holistic nature of PCA and ICA basis images. This parts based representation is a result of imposing nonnegativity onto the analysis. Since negative coefficients and basis elements are not allowed, the reconstruction is forced to be additive[13].

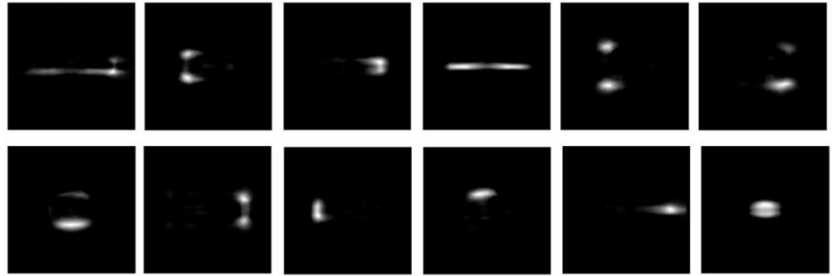

Figure 6. NMF basis for depth images

## 4. MATCHING

### 4.1 Axis rotations and reflections

One problematic issue with the CPCA normalization is the ambiguity of axis ordering and reflections. Most of the misalignment errors are due to inconsistent within-class axis orderings and orientations given by the normalization procedure. We resolve the axis ordering and reflection ambiguities by considering all 48 possible versions of the model, which we call ARR versions. We store the descriptors corresponding to each of these ARR versions for the models in the database.

Let $\Gamma = \{I_1, I_2, \ldots, I_{n_v}\}$ be the captured depth images of a model, where $n_v$ is the number of depth images. Let $\Gamma_r = \{I_{r1}, I_{r2}, \ldots, I_{rn_v}\}$ be the captured depth images of the same model with its axes re-labeled and reflected. For the geodesic sphere-based view sampling, there is a map from $\Gamma$ to $\Gamma_r$, which involves re-ordering of the depth images and rotating and reflecting the re-ordered depth images. We form the sets $\Gamma_1, \Gamma_2, \ldots \Gamma_{48}$ for the 48 different versions of a model (which we call ARR versions) through the 48 pre-defined mappings.

### 4.2 Pre-processing the database

Let the database consists of $D$ models, and let a database model indicated as $m_d$. The pre-processing steps applied to the database model $m_d$ are as follows:

1) <u>CPCA normalization:</u> Apply CPCA normalization to $m_d$ and calculate the eigenvalue ratios $a_1$ and $a_3$ of the model.

2) <u>Categorization and view sampling</u>: If $\sqrt{a_1^2 + a_3^2}$ is smaller then or equal to the categorization threshold $t_c$, then categorize the model as "elongated" and capture depth images from the 6 canonical directions, resulting in the image set $\Gamma = \{I_1, I_2, ..., I_6\}$ (as in Figure 3).

If $\sqrt{a_1^2 + a_3^2}$ is larger then the categorization threshold $t_c$, then categorize the model as "spherical" and capture depth images from the 18 vertices of the geodesic sphere, resulting in the image set $\Gamma = \{I_1, I_2, ..., I_{18}\}$ (as in Figure 3). Note that six images out of these 18 images correspond to the views seen from the six canonical directions. Indicate that subset as $\Gamma'$.

3) <u>Feature extraction</u>: Project each image onto the subspace (PCA, ICA or NMF), and obtain the set of feature vectors $F = \{\mathbf{f}_1, \mathbf{f}_2, ...\mathbf{f}_6\}$ if the model is "elongated". If the model is spherical, obtain $F = \{\mathbf{f}_1, \mathbf{f}_2, ...\mathbf{f}_{18}\}$. A subset of this set of feature vectors is indicated as $F'$.

4) <u>Axis rotations and reflections</u>: Re-arrange the images in $\Gamma$ in 48 different ways to obtain 48 sets of images $\Gamma_1, \Gamma_2, ..., \Gamma_{48}$ so that each set corresponds to an ARR version of the model (Section 4.1.). Apply the Feature extraction step to all the image sets to obtain 48 sets of feature vectors $F_1, F_2, ..., F_{48}$.

**4.3   Pre-processing the query model and matching**

Let the query model indicated as $m_q$. We apply the first three preprocessing steps described in 4.2. to the query model. If the query model is categorized as "elongated" we have $F^q = \{\mathbf{f}_1^q, \mathbf{f}_2^q, ...\mathbf{f}_6^q\}$. If it is categorized as "spherical" we have two sets of feature vectors as $F^q = \{\mathbf{f}_1^q, \mathbf{f}_2^q, ...\mathbf{f}_{18}^q\}$ and $F'^q = \{\mathbf{f}_1'^q, \mathbf{f}_2'^q, ...\mathbf{f}_6'^q\}$. We don't apply "axis rotations and reflections" step to the query model.

Table 1: Distance calculation according to the four cases of categorization of the query and database models.

|  | Query model is "elongated" | Query model is "spherical" |
|---|---|---|
| Database model is "elongated" | $n_v = 6$<br>$DIST(m_q, m_d) = \min_{r=1,2,...,48}\{dist(F^q, F_r^d)\}$<br>$dist(F^q, F_r^d) = \frac{1}{6}\sum_{i=1}^{6}\|\mathbf{f}_i^q - \mathbf{f}_{ir}^d\|$ | $n_v = 6$,<br>Use the subset $F'^q$ for comparison.<br>$DIST(m_q, m_d) = \min_{r=1,2,...,48}\{dist(F'^q, F_r^d)\}$<br>$dist(F'^q, F_r^d) = \frac{1}{6}\sum_{i=1}^{6}\|\mathbf{f}_i'^q - \mathbf{f}_{ir}^d\|$ |
| Database model is "spherical" | $n_v = 6$,<br>Use the subset $F'^d$ for comparison.<br>$DIST(m_q, m_d) = \min_{r=1,2,...,48}\{dist(F^q, F_r'^d)\}$<br>$dist(F^q, F_r'^d) = \frac{1}{6}\sum_{i=1}^{6}\|\mathbf{f}_i^q - \mathbf{f}_{ir}'^d\|$ | $n_v = 18$,<br>$DIST(m_q, m_d) = \min_{r=1,2,...,48}\{dist(F^q, F_r^d)\}$<br>$dist(F^q, F_r^d) = \frac{1}{18}\sum_{i=1}^{18}\|\mathbf{f}_i^q - \mathbf{f}_{ir}^d\|$ |

Let the distance between the query model $m_q$ and a database model $m_d$ be indicated as $DIST(m_q, m_d)$. If the eigenvalue ratios of the two models are not close enough, then $m_d$ is filtered out. If $\sqrt{(a_1^q - a_1^d)^2 + (a_3^q - a_3^d)^2}$ is higher than the filtering threshold $t_f$, the models are assumed to be completely different. The distance is returned as $DIST(m_q, m_d) \rightarrow \infty$, and the database model is cast at the end of the rank list.

If the database model $m_d$ is not filtered out the distance is calculated as follows:

$$DIST(m_q, m_d) = \min_{r=1,2,\ldots,48}\left\{dist(F^q, F_r^d)\right\}, \text{ where } dist(F^q, F_r^d) = \frac{1}{n_v}\sum_{i=1}^{n_v}\left\|\mathbf{f}_i^q - \mathbf{f}_{ir}^d\right\|.$$ $\|\bullet\|$ refers to $L_2$ norm. The variable $n_v$ indicates the number of depth images and it depends on the categorization of the query and database models. The distance calculation for four cases are shown in Table 1.

## 5. EXPERIMENTAL RESULTS

We conducted our experiments on the database of Princeton Shape Benchmark[11]. The database consists of a training set with 907 models in 90 classes and a test set with 907 models in 92 classes. We used nearest neighbor (NN), first tier (FT), second tier (ST), and discounted cumulative gain (DCG) as measures of retrieval performance. The definitions and implications of these measures can be found in the work of Shilane et al.[11].

### 5.1 View-based subspaces without categorization

Table 2 shows the results of view-based subspace approach on PSB test set without incorporating categorization and filtering of the objects with respect to eigenvalue ratios. In this case, the query and database models are rendered with a constant number of views (either 6 or 18). We set the filtering threshold to a high value, so that a query object is compared to all the objects in the database regardless of their eigenvalue ratios. The PCA, ICA and NMF basis views are obtained using the PSB training set, and the PSB test set is used for evaluation. The dimension in Table 2 indicates the number of basis views of each subspace, hence the number of the coefficients to describe one view.

First observation from Table 2 is that PCA approach gives significantly better results than ICA and NMF. We conjecture that PCA is more resilient to localization and alignment errors, on the other hand ICA and NMF techniques require good intra-class alignment. The views of an object greatly depends on the pose normalization; small errors in translation, scale and rotation normalization may result in significant mismatches in the views. For the rest of the experiments described in Sections 5.2 and 5.3 we use PCA-based features do describe the depth images.

Second observation is that increasing the number of views per object from 6 to 18 results in a non-negligible performance gain. However, this increase also brings greater computational cost, higher storage and longer search time. In Sections 5.2 and 5.3, we experimentally demonstrate that, in order to achieve that gain, we do not need to compare the query model to all the models in the database and we do not need to render 18 views from all the objects in the database but some of them.

Table 2. Retrieval performances of subspace methods without filtering and categorization with respect to eigenvalue ratios.

| Number of views | Subspace | Dimension | NN (%) | FT (%) | ST (%) | DCG (%) |
|---|---|---|---|---|---|---|
| 6 | PCA | 40 | 65.3 | 38.2 | 48.3 | 65.1 |
|   | ICA | 20 | 63.9 | 35.1 | 44.6 | 62.3 |
|   | NMF | 30 | 62.0 | 33.2 | 42.8 | 61.4 |
| 18 | PCA | 40 | **67.7** | **39.3** | **49.7** | **66.5** |
|   | ICA | 20 | 64.5 | 38.4 | 49.2 | 64.6 |
|   | NMF | 30 | 65.3 | 35.5 | 45.4 | 63.5 |

## 5.2 Filtering with respect to eigenvalue ratios

Figure 7 shows object samples from PSB arranged with respect to their eigenvalue ratios, $a_1$ and $a_3$, which are derived from the CPCA normalization. One can observer from Figure 7 that objects of the same class have close values of $a_1$ and $a_3$. For certain classes of objects, such as "standing" humans, cars, airplanes, or tires there is a limit for the variability of the eigenvalue ratios. These are objects where deformations and large body articulations are limited.

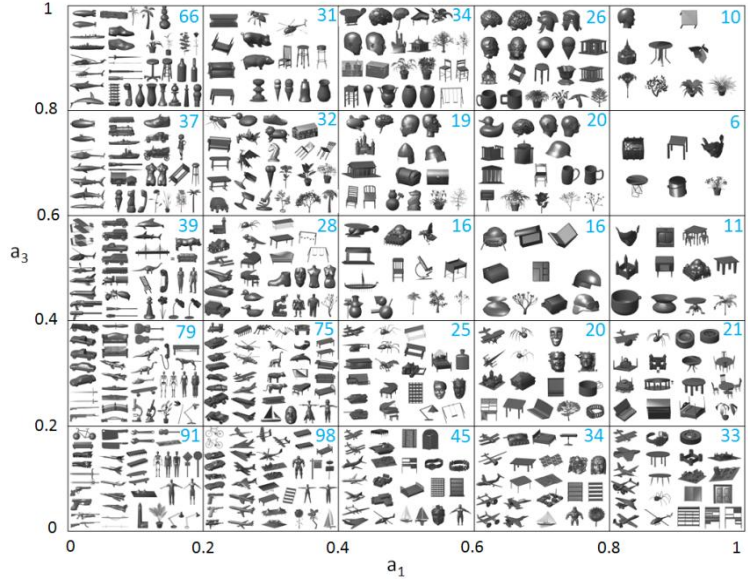

Figure 7. Object samples from PSB partitioned with respect to their eigenvalue ratios, $a_1$ and $a_3$.

As explained in Section 4.3, when a query model is introduced to the system, we calculate the distance between the eigenvalue ratios of the query model and each model in the database. If the distance is higher than a threshold, $t_f$, the dissimilarity between the query model and database model is set to infinity and the database model is cast at the end of the retrieval list. For the specific database of PSB we have searched for the optimum threshold to filter out the database models so that the retrieval performance is not affected by this pre-filtering process.

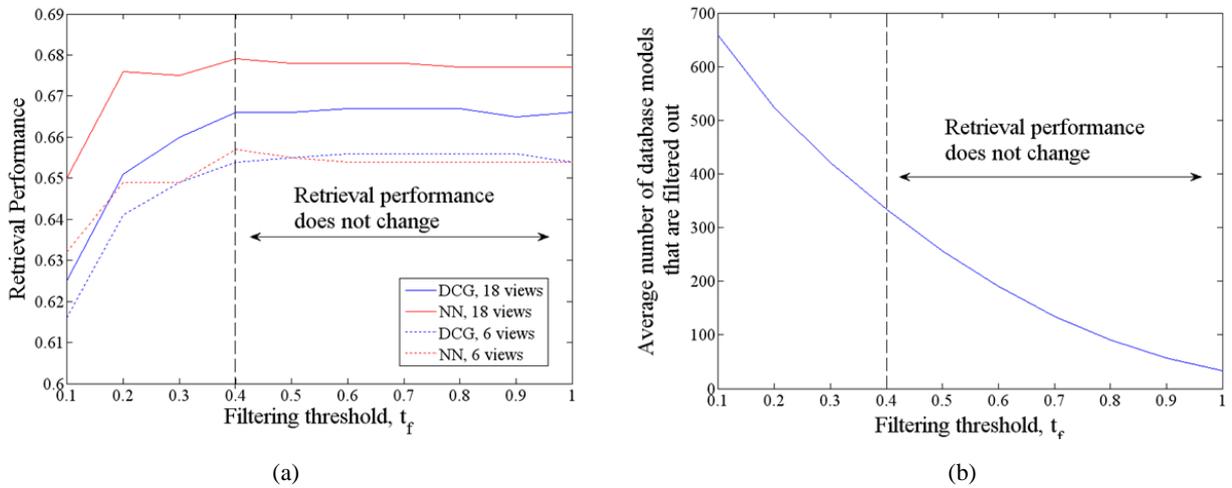

(a)    (b)

Figure 8. (a) Retrieval performance w.r.to the filtering threshold, $t_f$ for PSB test set. (b) Average number of database models that are filtered out w.r.to the filtering threshold, $t_f$.

Figure 8-(a) shows the Nearest Neighbor (NN) and Discounted Cumulative Gain (DCG) plotted with respect to varying values of $t_f$. The experiment is conducted such that all the models have the same number of views; either 6 or 18. From Figure 8-(a), we can observe that after a filtering threshold of 0.4, the retrieval performance does not change. This is equivalent to state that, for this particular database, there is no use to fully evaluate the dissimilarity between two models through the whole feature comparison process if their eigenvalue ratios are not close enough. Figure 8-(b) gives the average number of database models that are filtered out with respect to varying values of $t_f$. For $t_f = 0.4$, 333 database models out of 906 (one third of the database models) are filtered, hence are not matched fully to the query model. This is a very significant computational gain. For the experiments in the proceeding section, we set the filtering threshold to 0.4.

## 5.3 Categorization and variable number of views

As we suggested in Section 4.3, we do not need to obtain equal number of views from the query and database models in order to achieve a good retrieval performance. Instead, we categorized a model as "elongated" if the eigenvalue ratios are smaller than a categorization threshold, $t_c$; otherwise, we categorized it as "spherical". For the "elongated" models, we extracted and described only six views, for the "spherical" ones, 18 views were used. The matching between "elongated" and "spherical" models were performed as described in Section 4.3.

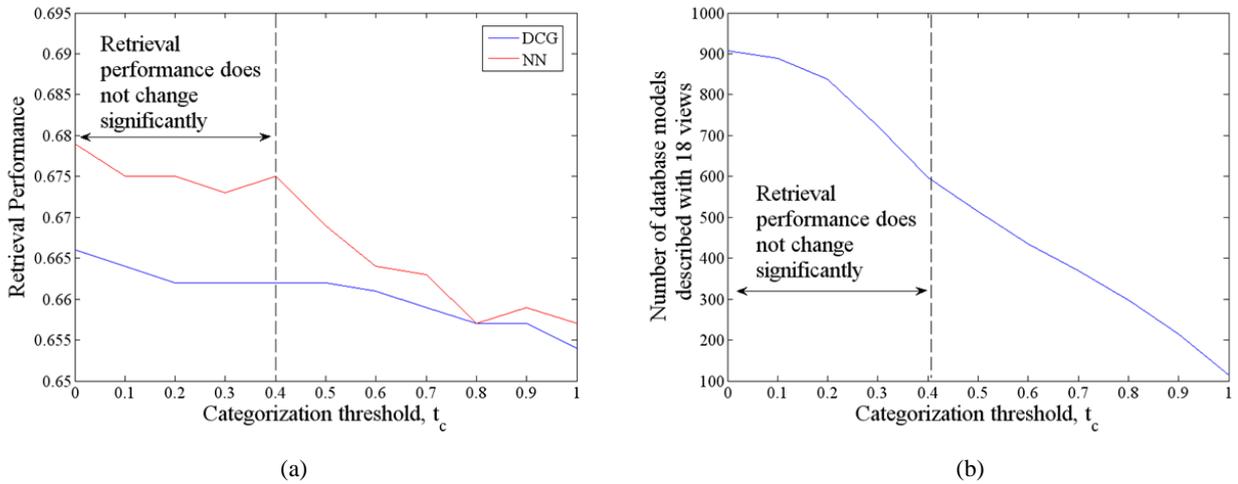

(a) (b)

Figure 9. (a) Retrieval performance w.r.to the categorization threshold, $t_c$ for PSB test set. (b) Number of database models that are described with 18 views w.r.to the categorization threshold, $t_c$.

Table 3. Retrieval performance and number of models categorized as "elongated" or "spherical".

| Categorization threshold $t_c$ | # "elongated" models (6 views only) | # "spherical" models (18 views) | NN (%) | FT (%) | ST (%) | DCG (%) |
|---|---|---|---|---|---|---|
| 0 | 0 | 906 | 67.9 | 39.4 | 49.6 | 66.6 |
| 0.4 | 311 | 595 | 67.5 | 39.5 | 49.6 | 66.2 |
| 1 | 793 | 113 | 65.7 | 38.6 | 48.9 | 65.4 |
| 1.37 | 906 | 0 | 65.6 | 38.2 | 48.3 | 65.1 |

Figure 9-(a) gives the retrieval performance in terms of NN and DCG with respect to varying values of the categorization threshold. Up to a threshold value of 0.4, the performance does not change significantly. However, as can be observed

from Figure 9-(b), at a threshold of 0.4, one third of the models in the database are described with only six views. Thus we obtain a similar retrieval performance while reducing the amount of storage and matching time significantly. Table 3 gives details of the retrieval performance, number of models categorized as "elongated" or "spherical" with respect to four different categorization thresholds.

### 5.4 Comparison with other view-based methods

In Table 4, we give a comparison of our method with some other view-based 3D model retrieval methods based on the results on PSB test set. We have used PCA subspace for feature extraction and we have set the filtering and categorization thresholds to 0.4. The reader can find the details of the mentioned algorithms in the corresponding references. The best performance is achieved by the method MDLA-DPD, and then follows our method. We believe that the other methods referenced in Table 4 would also benefit from the filtering and categorization processes we have proposed in this paper. These two processes would greatly reduce the computational cost by limiting the number of database models and views that are needed to be fully compared to the query model and its views.

Table 4. Comparison of our method with other view-based 3D model retrieval methods.

| Method | Number of views | NN (%) | FT (%) | ST (%) | DCG (%) |
|---|---|---|---|---|---|
| MDLA-DPD[7] | 20 | 68.8 | 43.6 | 54.2 | 67.8 |
| **Our method** | **Varying (6 or 18)** | **67.5** | **39.5** | **49.6** | **66.2** |
| DLA-DPD[7] | 6 | 66.7 | 39.5 | 50.2 | 65.3 |
| LFD[1] | 10 | 65.7 | 38.0 | 48.7 | 64.3 |
| AVC[4] | Varying (average 23) | 60.6 | 33.2 | 44.3 | 60.2 |

## 6. CONCLUSION

In this paper, we proposed a view-based 3D model retrieval system based on feature extraction via subspace methods. We compared three different subspace-based techniques, PCA, ICA and NMF, for feature extraction from the views of the 3D models. We have concluded that PCA-based feature extraction scheme outperformed ICA and NMF-based schemes for PSB test set. We plan to improve ICA and NMF-based schemes via working on the intra-class alignment mismatches of the extracted depth images.

In addition to the examining of view subspaces, we proposed two techniques to reduce the computational cost of the system: First is the pre-filtering of database models with respect to the distance between the eigenvalue ratios of the database and query models. The second is the rough categorization of the models as "elongated" and "spherical", and extracting different number of views from the models of each category. We have demonstrated that the retrieval performance is not affected by introducing these two techniques to the system and they yielded a significant drop in the computational cost.

**REFERENCES**

1. Chen, D.Y., Tian, X.P., Shen, Y.T. and Ouhyoung, M. "On visual similarity based 3D model retrieval," Computer Graphics Forum 22(3), 223–232 (2003).
2. Vranic, D., [3D Model Retrieval], Ph.D. Thesis, University of Leipzig (2004).
3. Furuya, T., Ohbuchi, R., "Dense Sampling and Fast Encoding for 3D Model Retrieval Using Bag-of-Visual Features," Proc. CIVR (2009).

4. Ansary, T.F., Daoudi, M. and Vandeborre, J-P., "3D model retrieval based on adaptive views clustering," Proc. ICAPR (2005).
5. Mahmoudi, S. and Daoudi, M, "3D models retrieval by using characteristic views," Proc. ICPR'02, 11-15 (2002).
6. Chaouch, M. and Verroust-Blondet, A., "Enhanced 2D/3D approaches based on relevance index for 3D-shape retrieval," Proc. SMI (2006).
7. Chaouch, M. and Verroust-Blondet, A., "A new descriptor for 2D Depth Image Indexing and 3D Model Retrieval," Proc. ICIP (2007).
8. Murase, H. and Nayar, S. K., "Visual learning and recognition of 3-D objects from appearance," Int. J. Comput. Vision 14(1), 5-24 (1995).
9. Vranic, D., Saupe, D. and Richter, J., "Tools for 3D-object retrieval: Karhunen-Loeve transform and spherical harmonics," Multimedia Signal Processing Workshop (2001).
10. Lian, Z., Rosin, P.L. and Sun, X., "Rectilinearity of 3D meshes," Int. J. Comput. Vision (in press).
11. Shilane, P., Min, P., Kazhdan, M. and Funkhouser, T., "The Princeton Shape Benchmark," Proc. SMI, 167-178 (2004).
12. Hyvärinen, A. and Oja, E., "Independent Component Analysis: Algorithms and Applications," Neural Networks, 13(4-5), 411-430 (2000).
13. Lee, D. D. and Seung, H. S., "Learning the parts of objects by nonnegative matrix factorization," Nature 401, 788–791 (1999).